\documentclass[letterpaper, 10 pt, conference]{IEEEtran}
\IEEEoverridecommandlockouts
\usepackage{cite}
\usepackage{amsmath,amssymb,amsfonts}
\usepackage{graphicx}
\usepackage{textcomp}
\usepackage{xcolor}
\usepackage{caption}
\usepackage{booktabs}
\usepackage{makecell}
\usepackage{multirow}
\usepackage{soul}

\usepackage{float}
\usepackage{algorithm}
\usepackage{algpseudocode}

\makeatletter
\let\NAT@parse\undefined
\makeatother
\usepackage{hyperref}

\usepackage{lipsum}  

\usepackage{subcaption}
\graphicspath{ {./figures/} }

\DeclareMathOperator*{\argmax}{argmax} 

\def\BibTeX{{\rm B\kern-.05em{\sc i\kern-.025em b}\kern-.08em
    T\kern-.1667em\lower.7ex\hbox{E}\kern-.125emX}}
\begin{document}

\title{Exploring Unstructured Environments using Minimal Sensing on Cooperative Nano-Drones}

\author{
    Pedro Arias-Perez$^{1}$, Alvika Gautam$^{2}$, Miguel Fernandez-Cortizas$^{1}$, \\ David Perez-Saura$^{1}$, Srikanth Saripalli$^{2}$ and Pascual Campoy$^{1}$ 
    \thanks{$^{1}$is with the Computer Vision and Aerial Robotics Group at Universidad Politécnica de Madrid, Spain (CVAR-UPM) at the Centre for Automation and Robotics C.A.R. (UPM-CSIC). \tt\small{\{pedro.ariasp, miguel.fernandez.cortizas, david.perez.saura, pascual.campoy\}@upm.es}}
    \thanks{$^{2}$is with the Department of Mechanical Engineering, Texas A\&M University, College Station, TX, USA \tt\small{\{alvikag, ssaripalli\}@tamu.edu}}
    \thanks{*This work was partially funded by FPU20/07198 of the Spanish Ministry for Universities and by the project INSERTION ref. ID2021-127648OBC32, ``UAV Perception, Control and Operation in Harsh Environments'', funded by the Spanish Ministry of Science and Innovation under the program ``Projects for Knowledge Generating''.}
    \thanks{For the purpose of Open Access, and in fulfillment of the obligations arising from the grant agreement, the author has applied a Creative Commons Attribution 4.0 International (CC BY 4.0) license to any Author Accepted Manuscript version arising from this submission.}
}
\maketitle
\begin{abstract}

Recent advances have improved autonomous navigation and mapping under payload constraints, but current multi-robot inspection algorithms are unsuitable for nano-drones due to their need for heavy sensors and high computational resources. To address these challenges, we introduce \emph{ExploreBug}, a novel hybrid frontier range bug algorithm designed to handle limited sensing capabilities for a swarm of nano-drones. This system includes three primary components: a mapping subsystem, an exploration subsystem, and a navigation subsystem. Additionally, an intra-swarm collision avoidance system is integrated to prevent collisions between drones. We validate the efficacy of our approach through extensive simulations and real-world exploration experiments involving up to seven drones in simulations and three in real-world settings, across various obstacle configurations and with a maximum navigation speed of 0.75 m/s. Our tests demonstrate that the algorithm efficiently completes exploration tasks, even with minimal sensing, across different swarm sizes and obstacle densities. Furthermore, our frontier allocation heuristic ensures an equal distribution of explored areas and paths traveled by each drone in the swarm. We publicly release the source code of the proposed system to foster further developments in mapping and exploration using autonomous nano drones. 

Video: \url{https://vimeo.com/cvarupm/cf-exploration}

Code: \url{https://github.com/pariaspe/cf_exploration}

\end{abstract}

\section{Introduction} \label{section-introduction}


Exploration and mapping of unknown unstructured environments is a crucial area of research in the field of mobile robotics. As the technology of nano-sized drones advances \cite{palossi2017target}\cite{zhang2015autonomous}\cite{jaisinghani2023iot}, their utility in both indoor and outdoor autonomous tasks has gained considerable prominence. A nano-sized drone typically weighs $\thicksim$50 g or less and is the size of a palm. These drones are particularly well-suited for potentially hazardous indoor exploration missions due to their small size, which ensures safety around humans, and their agility and speed, which are essential for effective exploration. Moreover, their low cost contributes to their feasibility for widespread use as they are easily replaceable.

\begin{figure}[!ht]
    \centering
    \includegraphics[width=0.49\textwidth]{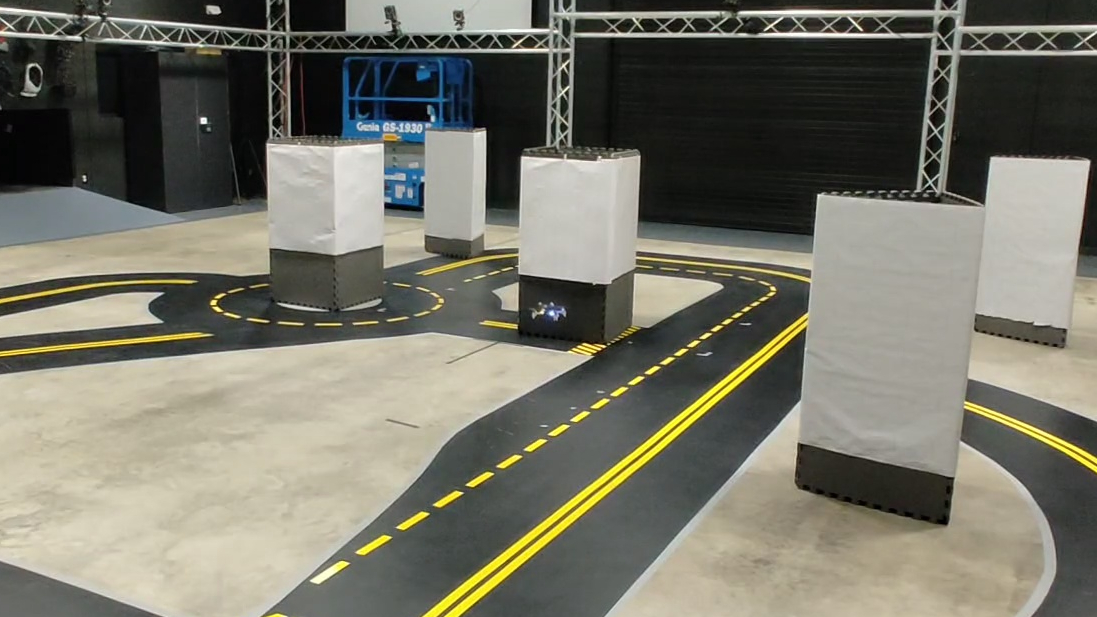}
    \caption{View of one nano-drone performing a exploration of an unknown unstructured environment.}
    \label{fig:portrait}
    \vspace{-0.2cm}
\end{figure}

Building on the capabilities of individual nano-drones, the use of drone swarms enhances these operations significantly. Swarms leverage the collective speed and maneuverability of multiple drones, which theoretical and empirical research confirms can accomplish tasks more efficiently and rapidly than a single drone \cite{guzzoni1997many}. Moreover, the strategic deployment of multiple drones introduces a level of redundancy, thereby enhancing the overall system's fault tolerance, compared to a single drone. Additionally, cooperating robots can significantly reduce sensor uncertainty thanks to the information overlap between different robots. 
 

Several authors in the literature have shown promising progress in robotic exploration \cite{burgard2005coordinated, zhu2015frontier, oleynikova2018safe}. Available algorithms strongly rely on complex sensors, as depth cameras or 3D lidars, for obstacle avoidance and mapping. However, current nano-drones can not adopt these algorithms as the restricted payload resources for on-board sensors and computational power. The available payload in nano-drones is in the order of grams. Therefore, it is only possible to mount light and low-power sensors that can only provide sparse and noisy measurements. Consequently, the minimal sensing capacity introduces a challenging exploration problem.

In this work, we present \emph{ExploreBug}, a coordination algorithm for a group of nano-drones to perform an efficient full exploration. Our method aims to address the challenge of minimal sensing exploration. Against latest contributions in the topic, where authors focus on exploration time and speed, we pursue to grow smaller not stronger. Our aim is to evaluate similar techniques on smaller platforms, how far we can push the limits and evaluate the trade-offs accordingly.


Concisely, the main contributions of this work are as follows:
\begin{enumerate}
    \item Design of a novel exploration algorithm that effectively utilizes minimal sensing capabilities (four single-beam range sensors) tailored specifically for nano-drones.
    \item Extensive experimental evaluation of the proposed approach in simulated and real environments, demonstrating promising results. 
    \item Finally, we contribute by making the code of the proposed system open-source enabling researchers to evaluate and develop upon our framework. 
\end{enumerate}

\section{Related Work} \label{section-related-work}

\subsection{Multi-robot Exploration} 
Autonomous exploration of unknown environments has been a common robotics research topic over the last two decades. One of the most popular approaches is a frontier-based exploration technique, first introduced in \cite{yamauchi1997frontier}. The main concept behind frontier-based exploration is moving to the boundary where explored space meets uncharted territory to maximize new information about the world. By constantly moving to these new frontiers, a robot can expand its map into unexplored areas until the entire environment is fully mapped.
Burgard et al. present one of the first frontier-based exploration solution for a team of robots \cite{burgard2005coordinated}. Their approach chooses the appropriate target points for the individual robots so that simultaneously takes into account the cost of reaching a target point and its utility, which is reduced whenever the target point is assigned to a robot.
In \cite{mannucci2018autonomous}, Mannucci et al. propose a 3D exploration strategy based on a combination of local and global information for a team of ﬂying robots with constrained payload. 
Motivated by the insufficient exploration rate, Zhou et al. provide FUEL \cite{zhou2021fuel}, a hierarchical framework that can support fast drone exploration in complex unknown environments. Later, they extend their prior work for a fleet of
decentralized drones, suggesting a new system named RACER \cite{zhou2023racer}. They decompose the unknown space into hgrid and distribute the task units among quadrotors by a pairwise interaction, assuming asynchronous and unreliable communication.
Hui et al. consider a limited communication bandwidth between the multi-drone system. Their decentralized exploration planning framework, named DDPM \cite{hui2023dppm}, uses a lightweight information structure which enables exploration
with little overlap. Each drone maintains a topological graph to save the information structure, serving as a high-level understanding and abstraction of the environment.
Bartolomei et al. suggest a higher-level reasoning for multi-robot missions \cite{bartolomei2023fast}. Each individual drone has the ability to alternate between various execution modes. This balance involves a cautious exploration of yet completely unknown regions and more aggressive exploration of smaller, unfamiliar areas within the space.


Nevertheless, despite the progress in the topic in recent years, there are big limitations to apply these solutions on nano-drones. All systems previously cited require heavy sensors, as depth cameras or 3d lidars, and high computational power of their algorithms. Resource restrictions on nano-drones rule out the currently available solutions for frontier-based exploration.

\subsection{Nano-Drones \& Minimal Sensing}
Recently, several studies have shown enormous improvement in autonomous navigation and mapping \cite{zhou2022efficient, niculescu2023nanoslam}, both key components to fulfill an exploration. As a result of the extreme payload constraints of nano-sized drones, previous works have sought alternative exploration strategies. \\
A promising bug exploration algorithm was introduced by McGuire et al. \cite{mcguire2019minimal}. This algorithm empowered a swarm of nano quadcopters to effectively navigate through unfamiliar and cluttered environments, ultimately returning to their initial starting location.
In \cite{duisterhof2021sniffy}, Duisterhof et al. propose Sniffy Bug, a bug-inspired autonomous swarm to seek gas leaks in unknown environments. Their exploration strategy rely in the observed gas concentrations by the swarm, which indicates search directions, while the bug navigation is performed in two stages, line following and wall following.
In \cite{lamberti2023bio}, Lamberti et al. present a bio-inspired autonomous exploration for object detection. The authors compare four different navigation policies from random to scan spinning, showing the pseudo-random policy the best result.
Pourjabar et al. suggest a system design to achieve robust autonomous exploration capabilities for a swarm of nano-drones in \cite{pourjabar2023multi}. Their multi-sensory approach combines lightweight single-beam laser ranging, low resolution camera and an ultra-wide-band based ranging module.


However, none of these approaches exploit the benefits of the environment representation of frontier-based exploration methods. Motivated by this scarcity, in this work we present a minimal sensing frontier-based exploration technique capable to be carried by nano-drones. 

\section{Methodology} \label{section-methodology}
The overall problem considered in this work is to perform a full map exploration with a nano-drone swarm.
The main constraint of nano drones is their low capabilities on payload. Therefore, onboard sensors are limited which implies spare sensing and mapping. We propose an exploration pipeline that can deal with minimal sensing. This strategy allows a full exploration for nano-drones.
Due to its limited payload, in our system each drone only carries four single-beam range sensors, pointing front, back, right, and left, and 4 meters maximum range. Because of the sensor used, the exploration performed is a 2D problem, so that the drones don't change their altitude during their mapping and exploration task. \\
Another relevant assumption is that agents can be localized in a common reference frame and they share an infinite-range communication channel between them.


\subsection{System Overview} 

\begin{figure}[h]
    \centering
    \includegraphics[width=0.49\textwidth]{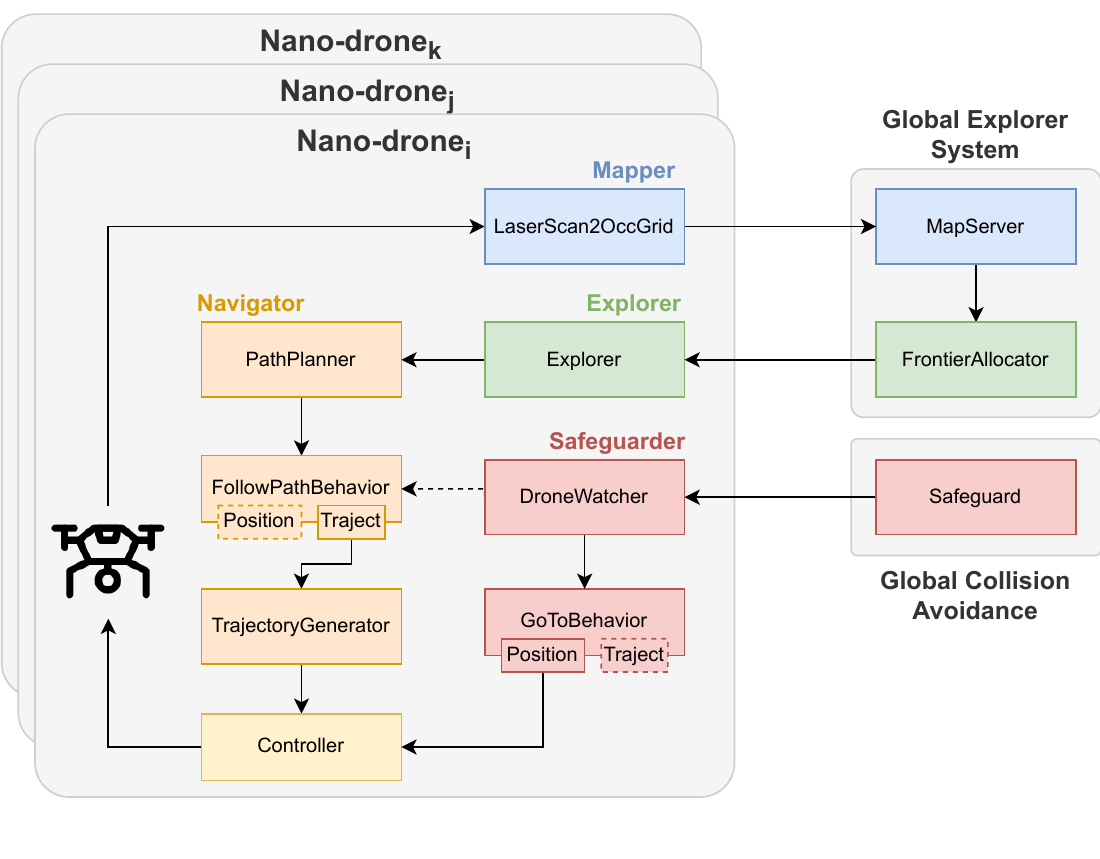}
    \caption{Overview of the pipeline. Blue boxes belong to the mapping sub-system, green boxes to the exploration sub-system, orange boxes to the navigator sub-system and red boxes to the intra-swarm collision avoidance sub-system.}
    \label{fig:pipeline}
    \vspace{-0.3cm}
\end{figure}


The system comprises of three main components: a mapping sub-system, an exploration sub-system, and a navigator sub-system, each corresponding to a specific phase of the mission. Fig. \ref{fig:pipeline} provides an overview of the pipeline. 

In the mapping phase, each agent utilizes readings from the multi-ranger sensor, shared among the swarm, to create a common map. Given that the sensor collects only 2D information about the environment, the map is encoded as an occupancy grid. 

The exploration phase consists of three steps: the exploration strategy, frontier generation, and frontier allocation. The exploration strategy employs a hybrid frontier range bug algorithm. Subsequently, a set of available frontiers is generated from the map, and based on the swarm's state, the most suitable frontier is allocated on demand to each agent. 

Finally, the navigator transmits motion references to the controller to reach the goal point. The navigation system's primary component is a local path planner, leveraging Aerostack2 \cite{aerostack2023}, a framework for developing autonomous aerial robotics systems. Aerostack2 facilitates the initiation, pausing, or cessation of high-level robotics abstractions, known as behaviors (e.g., go to way-point, hover, or take off), streamlining drone motion control.

Alongside these systems, an intra-swarm collision avoidance system operates to prevent collisions among drones. It tracks the position of each drone and, if a potential collision is detected, it temporarily halts one drone's navigator and initiates a safety maneuver to prevent the collision. Once the safety distance is reestablished, the navigator resumes its operation, ensuring the exploration continues safely.



\subsection{Explore Bug}

\begin{figure}[ht]
    \centering
    \includegraphics[width=0.35\textwidth]{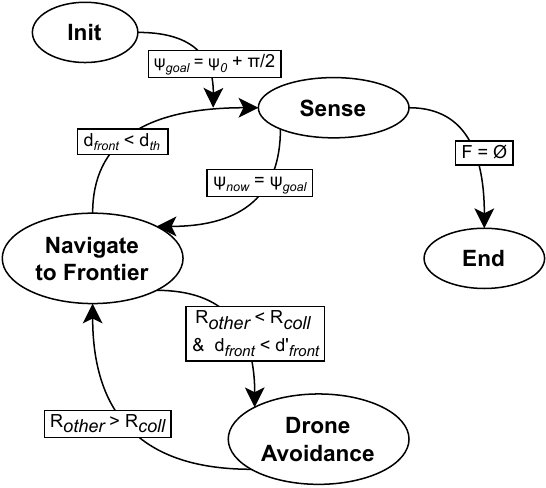}
    \caption{FSM of the ExploreBug algorithm.}
    \label{fig:fsm}
\end{figure}

The exploring strategy operates as a finite state machine (FSM), depicted in Fig. \ref{fig:fsm}, where each drone begins at the ``Init'' state. The approach is similar to \textit{CautiousBug} \cite{magid2004cautiousbug} or \textit{Rotate-and-measure} \cite{lamberti2023bio}, where a conservative spiral search is performed before going to a specific location. This ensures a safer exploration strategy by thoroughly scanning the environment before navigation. During the ``Sense'' phase, the drone continuously collects data from the surroundings at a rate of 20 Hz. 

Frontiers, which represent potential next points for navigation, are identified over the generated map, leading to the ``Navigate to Frontier'' state. It should be noted that the term navigation implies the need of a map and can only be performed on the already explored surroundings. Frontier generation and allocation to drones is described in subsequent subsections in detail. Upon reaching the goal ($d_{front} < d_{th}$), the robot returns to the ``Sense'' state.

Due to the symmetry of the range-sensor, spinning consist of a $\frac{\pi}{2}$ radians turn only $\psi_{goal} = \psi_0 + \frac{\pi}{2}$. Rotating just occurs during the ``Sense'' state since navigation can be performed without yaw control. Our cautious algorithm ensures to fly under already known environment only, so the drone will not fly over an obstacle.

During the navigation state, if two or more drones are within a safety margin distance $R_{other} < R_{coll}$, the one that's closest to its goal ($d_{front} < d'_{front}$) transitions into the ``Drone Avoidance'' state (that is, change in altitude and hover to avoid collision), while the other drone continues to fly without any avoidance maneuver. This ensures a quicker return to the navigation state once the safety distance is restored.
Finally, when no frontiers are left, the exploration is consider as finished and the algorithm enters into ``End'' state, which is accessible from any other state within the FSM.

\begin{figure*}[th!]
    \centering
    \begin{subfigure}[b]{0.245\textwidth}
        \centering
        \includegraphics[width=\textwidth]{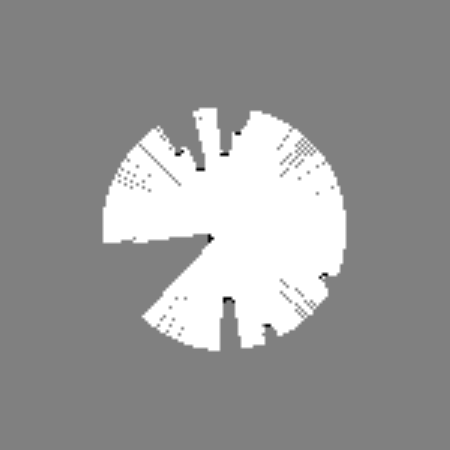}
        \caption{Map}
        \label{fig:map_original}
    \end{subfigure}
    \hfill
    \begin{subfigure}[b]{0.245\textwidth}
        \centering
        \includegraphics[width=\textwidth]{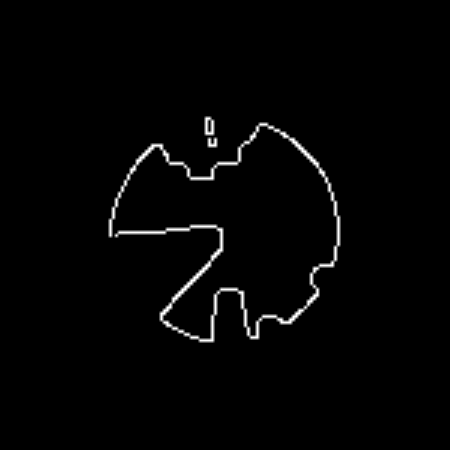}
        \caption{Edges}
        \label{fig:map_edges}
    \end{subfigure}
    \hfill
    \begin{subfigure}[b]{0.245\textwidth}
        \centering
        \includegraphics[width=\textwidth]{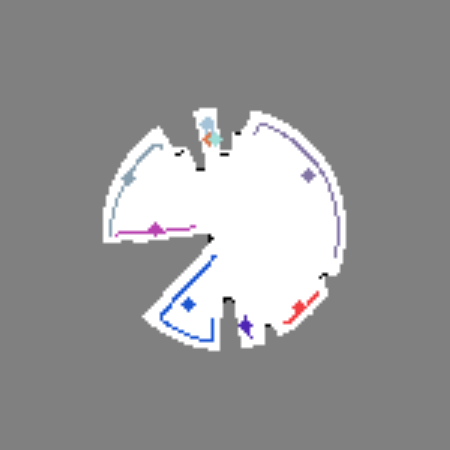}
        \caption{Frontiers clustered}
        \label{fig:map_cluster}
    \end{subfigure}
    \hfill
    \begin{subfigure}[b]{0.245\textwidth}
        \centering
        \includegraphics[width=\textwidth]{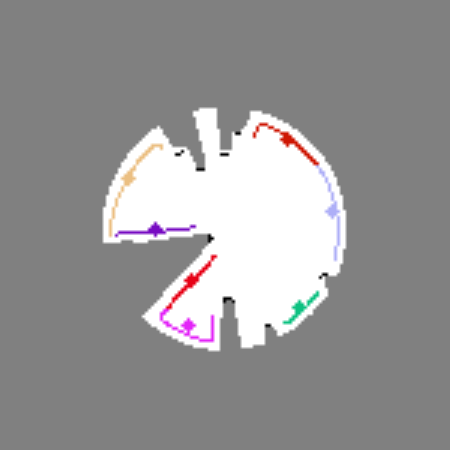}
        \caption{Frontiers filtered}
        \label{fig:map_filter}
    \end{subfigure}
    \caption{Frontier generation pipeline: (a) map, (b) edges extraction, (c) frontiers clustered and (d) frontiers filtered. $M_{free}$ are represented in white, $M_{obs}$ in black and $M_{unk}$ in gray. Each color pictures pixels and centroid from every frontier.}
    \label{fig:frontier_generation}
    \vspace{-0.2cm}
\end{figure*}

\subsection{Frontier Generation}
Given the sensor's nature, occupancy grid maps are selected to represent the environment. These maps maintain the probability of each grid cell being empty or occupied. Occupancy maps have been proven to be an efficient and powerful description of the surroundings incrementally updating the map based on the sensor readings \cite{moravec1988sensor}. 


Local occupancy grids are received from each drone and stored in a cell grid central map $M$. Cells can be unknown or have a certainty to be occupied/free, $M = M_{free} \cup M_{obs} \cup M_{unk}$. This occupation probability is updated based on the hit or miss of the sensor readings. The central map is used to obtain a set of frontiers $F$, where $F \subseteq M$. A frontier $f_i$ is formed by a group of consecutive cells in the grid and it is determined by a unique frontier descriptor. A descriptor is formed by the centroid $c_i$, orientation $\psi_i$ and area $A_i$ (or number of cells) of the frontier.

\begin{equation}
    \forall f_i \in F , \; f_i : \{c_i, \psi_i, A_i\}
\end{equation}

Frontiers $F$ are obtained from the relative complement of $M_{obs}$ in the edges detected over $M_{free}$, $canny(M_{free})$ (Eq. \ref{eq:frontiers}). After the extraction, frontiers are clustered based on pixel connectivity and filtered depending on their area. If a frontier is smaller than a lower threshold it is discarded, whether if it is bigger that an upper threshold it is split. The Algorithm \ref{alg:front_gen} describes frontier generation.

\begin{equation}
    F = canny(M_{free}) \cap M_{obs}
    \label{eq:frontiers}
\end{equation}

Fig. \ref{fig:frontier_generation} shows the frontier generation pipeline, from the global map $M$ (Fig. \ref{fig:map_original}) to the frontiers extracted $F$ (Fig. \ref{fig:map_filter}). Edges detected over the map $canny(M_{free})$ are presented in Fig. \ref{fig:map_edges}. Finally, Fig. \ref{fig:map_cluster} exhibits the frontiers clustered previous to the filtering. 

\begin{algorithm}
\caption{Frontier generation}\label{alg:front_gen}
\begin{algorithmic}
\Require $M$: current grid map
\State $ \bar{M}_{free} = M_{obs} \cup M_{unk} $
\State $ \bar{M}_{obs} = M_{free} \cup M_{unk} $
\State $ E = $ canny($\bar{M}_{free}$)
\State $ F = $ $\bar{M}_{obs} \cap E $
\ForAll{$ f_i \in F $}
    \If{$A_i < lower\_threshold $}
        \State remove $f_i$ from $F$
    \ElsIf{$A_i > upper\_threshold $}
        \State split $f_i$ \Comment{By pixel continuity}
    \EndIf
\EndFor
\end{algorithmic}
\end{algorithm}


 

\subsection{Frontier Allocator}
Goal choosing relies on the frontier candidates and the drone swarm. The optimization problem is defined in Equation \ref{f_cost}. Cost function depends on two terms. First minimizes the path to next frontier. We believe that the exploration rate is maximized during the ``Sense'' phase and not during the ``Navigation'' state since navigation occurs over already mapped area. Second term minimizes the overlap between agents, maximizing distance between them.

\begin{equation}
    G^{\ast} = \argmax_G \quad \frac{1}{X} \sum_{i}^{F} \sum_{j}^{X} d_{ij} - 2 \sum_{i}^{F} c_i
    \label{f_cost}
\end{equation}

where $G^{\ast}$ is the optimal frontier, $X$ is the set of drones, $F$ is the set of frontiers, $d_{ij}$ is the distance between the frontier allocated to drone $i$ and frontier $j$ and $c_i$ is the distance between frontier $i$ and the current drone. 

The proposed heuristic does not have any constraints related to the robot motion. Exploration algorithm trades safety against speed. Each nano-drone is already stopping when reaching the goal frontier to fulfill the scan action, so motion speed is not taken into account while allocating next frontier. Moreover, there is no need to change heading during navigation since path goes through known space and sensor symmetry.
\section{Experimental Validation} \label{section-experimental-validation}
To show the performance of our algorithm, we performed several experiments varying world size, obstacle density and swarm size. The experimental validation was done in both real and simulated environments.

\subsection{Experimental setup}
The nano-drones used to form the swarm in our experiments are Bitcraze Crazyflie 2.1 quadcopters, each weighing 27 grams with a diameter of 10 cm. These drones are equipped with a multi-ranger deck featuring five VL53L1x Time-of-Flight (ToF) sensors, capable of measuring distances up to 4 meters with millimeter precision at a frequency of 20 Hz. Additionally, each drone carries a motion capture marker deck, with the total drone payload, including the battery, being under 10 grams.

Our experiment scenarios replicated non-structured environments, using cylindrical poles as obstacles that imitate forest trees for both simulated and real environments (see Fig. \ref{fig:portrait}). The variety in our experiments stemmed from different parameters such as the density of these obstacles, the size of the drone swarm, and the initial placement of drones and obstacles' location within the testing area.

Simulations are conducted using Gazebo replicating Crazyflie characteristics. The quadcopter dynamics are simulated based on RotorS \cite{Furrer2016} using the ground truth of the drone. A multi-ranger deck imitation is placed onboard as the only payload.

The exploration algorithm leverages Aerostack2 \cite{aerostack2023} for navigation and control, enabling the use of a consistent algorithm across both simulated and real-world setups. However, there are slight variations in the Aerostack2 components used in each system. Specifically, the interfaces connecting the platforms to Aerostack2 differ and the state estimator plugins vary. Aerostack2 functions uniformly across both environments, providing a seamless transition from simulation to actual deployment.


\subsection{Exploration in Simulated Scenarios}
We benchmark the proposed exploration strategy in a Gazebo simulation. We test the algorithm using different swarm size, setting diverse starting points for the swarm, varying the obstacle density and world size. In all experiments, we set the navigation speed to $0.75 m/s$, the spin speed to $0.15$ $rad/s$, and the sensor range to $4m$. The remaining parameters are shown in Table \ref{tab:sim_conf}.

\begin{table}[h]
    \centering
    \begin{tabular}{c|c}
        \hline \hline
       \textbf{Parameter} & \textbf{Value} \\
        \hline
        Map resolution & 0.1 \\
        Sensor range & 4 \\
        Safety distance to obstacles & 0.4 \\
        Safety distance to other drones & 2.0 \\
        Navigation speed & 0.75 \\
        Reached distance threshold & 0.2 \\
        Spin speed & 0.15 \\
        Yaw threshold & 0.15 \\
        Minimum frontier size & 1.5 \\
        Maximum frontier size & 3.5 \\
        \hline \hline
    \end{tabular}
    \caption{Parameter configuration for simulation experiments.}
    \label{tab:sim_conf}
    \vspace{-0.2cm}
\end{table}

\begin{figure}[h]
    \centering
    \begin{subfigure}[b]{0.43\textwidth}
        \centering
        \includegraphics[width=\textwidth]{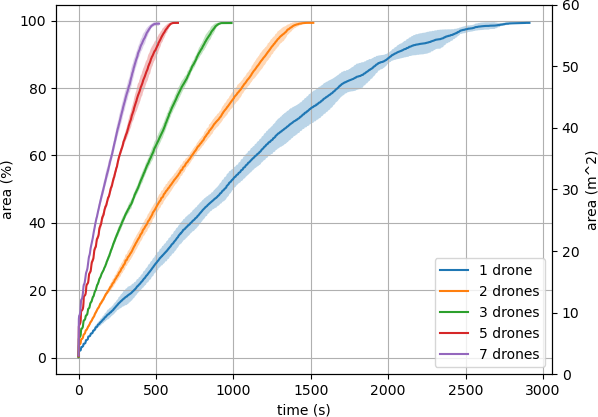}
        \caption{Average exploration rate for the 0.05 $obs/m^2$ density scenario. The shaded area shows the standard deviation.}
        \label{fig:area_sim_results}
    \end{subfigure}
    \par\bigskip
    \begin{subfigure}[b]{0.43\textwidth}
        \centering
        \includegraphics[width=\textwidth]{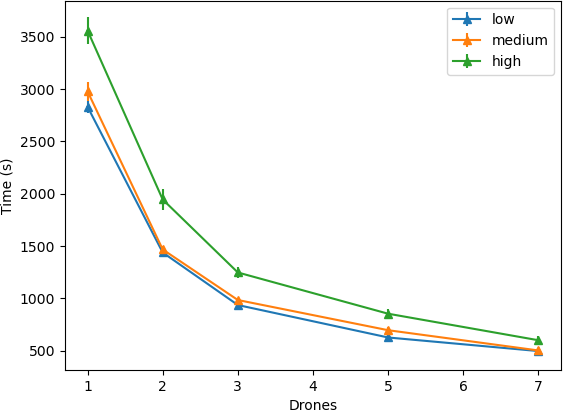}
        \caption{Exploration time per swarm size depending on the obstacle density. Low stands for 0.05 $obs/m^2$, medium for 0.1 $obs/m^2$ and high for 0.2 $obs/m^2$.}
        \label{fig:obs_dens_comp}
    \end{subfigure}
    \caption{Simulation experiment results up to seven drones.}
    \label{fig:sim_results}
    \vspace{-0.4cm}
\end{figure}

\begin{table*}[!h]
    \centering
    \begin{tabular}{cccccc}
        \hline \hline
         Obstacle density [$obs/m^2$] & \# of drones & Time [s] & Area [\%] & Path Length [m] & Overlap [\%] \\
        \hline
        \multirow{5}{1em}{0.05} & 1 & 2831.79 $\pm$ 59.62 & 99.37 $\pm$ 0.01 & 986.66 $\pm$ 24.51 & - \\
                                & 2 & 1438.00 $\pm$ 45.98 & 99.37 $\pm$ 0.02 & 972.4 $\pm$ 34.08 & 25.79 $\pm$ 8.67 \\
                                & 3 & 935.40 $\pm$ 24.79  & 99.32 $\pm$ 0.01 & \textbf{913.10 $\pm$ 23.65} & \textbf{16.88 $\pm$ 3.27} \\
                                & 5 & 624.98 $\pm$ 21.77 & \textbf{99.37 $\pm$ 0.02} & 920.92 $\pm$ 26.87 & 31.32 $\pm$ 5.17 \\
                                & 7 & \textbf{494.99 $\pm$ 22.17} & 99.14 $\pm$ 0.66 & 1037.30 $\pm$ 46.89 & 35.26 $\pm$ 5.06 \\
        \hline
        \multirow{5}{1em}{0.1} & 1  & 2979.00 $\pm$ 91.46  & 98.56 $\pm$ 0.06 & 1008.97 $\pm$ 37.59 & - \\
                               & 2  & 1467.00 $\pm$ 29.25 & 98.62 $\pm$ 0.04 & 998.47 $\pm$ 18.87 & 24.0 $\pm$ 6.16 \\
                               & 3  & 983.39 $\pm$ 23.43  & 98.62 $\pm$ 0.03 & \textbf{941.17 $\pm$ 22.53} & \textbf{17.85 $\pm$  2.52} \\
                               & 5  & 694.60 $\pm$ 27.77 & 98.63 $\pm$ 0.02 & 1008.51 $\pm$ 39.45  & 30.29 $\pm$ 5.45 \\
                               & 7  & \textbf{500.80 $\pm$ 20.02} & \textbf{98.67 $\pm$ 0.02} & 1126.36 $\pm$ 34.72 & 36.06 $\pm$ 3.44 \\
        \hline
        \multirow{5}{1em}{0.2} & 1  & 3560.80 $\pm$ 125.83 & 95.19 $\pm$ 1.13 & 1153.33 $\pm$ 19.85 & - \\
                               & 2  & 1944.99 $\pm$ 99.69 & 96.26 $\pm$ 0.62 & 1214.82 $\pm$ 86.44 & 26.42 $\pm$ 8.81 \\
                               & 3  & 1246.79 $\pm$ 49.86 & 95.02 $\pm$ 0.98 & \textbf{1153.14 $\pm$ 64.81} & \textbf{25.18 $\pm$ 8.35} \\
                               & 5  & 852.38 $\pm$ 49.70  & 96.28 $\pm$ 1.02 & 1265.07 $\pm$ 69.12 & 27.73 $\pm$ 3.01 \\
                               & 7  & \textbf{598.61 $\pm$ 37.37} & \textbf{96.54 $\pm$ 0.43} & 1288.79 $\pm$ 91.09 & 35.71 $\pm$ 3.99 \\
        \hline \hline
    \end{tabular}
    \caption{Results of the simulation experiments up to seven drones. We report the average and standard deviation for the completion time, percentage of area explored and total path traveled for the swarm. Best results are highlighted in bold. Lowest times correspond to the highest drones number, nonetheless shortest path lengths match with lowest overlap. Increasing the swarm size slightly increases the overlap.}
    \label{tab:sim_results}
    \vspace{-0.2cm}
\end{table*}

We assemble three simulation universes with different obstacle densities. Each scenario is 50x50m in size with an obstacle density of 0.05, 0.1, and 0.2 $obs/m^2$ respectively. We perform experiments with one, two, three, five, and seven drones for each proposed environment.

\begin{figure*}[!ht]
    \centering
    \begin{subfigure}[b]{0.32\textwidth}
        \centering
        \includegraphics[width=\textwidth]{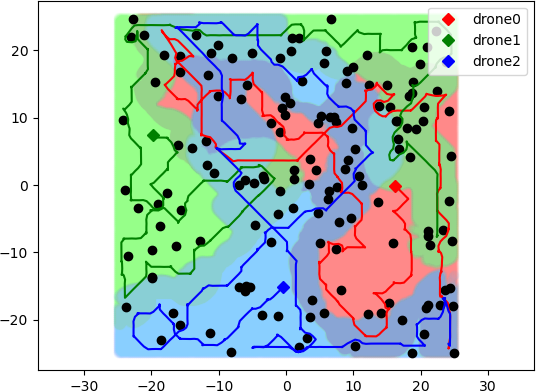}
        \caption{Nearest frontier. Time 1030$s$, path length 1026.1$m$ and overlap 34.48$\%$.}
        \label{fig:3_closest}
    \end{subfigure}
    \hfill
    \begin{subfigure}[b]{0.32\textwidth}
        \centering
        \includegraphics[width=\textwidth]{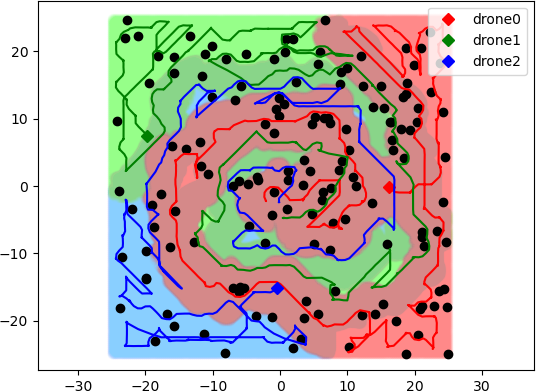}
        \caption{Maximum distance between drones. Time 1562$s$, path length 1329.73$m$, overlap 46.4$\%$.}
        \label{fig:3_max_dist}
    \end{subfigure}
    \hfill
    \begin{subfigure}[b]{0.32\textwidth}
        \centering
        \includegraphics[width=\textwidth]{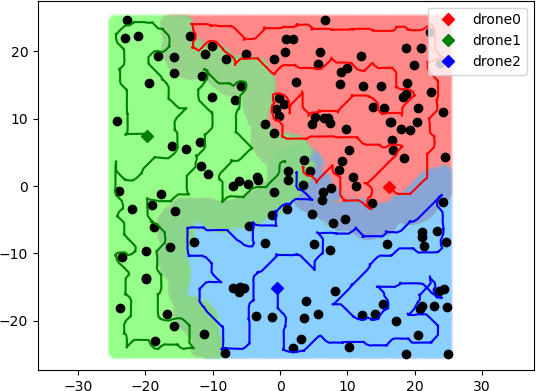}
        \caption{Ours. Time 942$s$, path length 917.06$m$ and overlap 14.18$\%$.}
        \label{fig:3_final_heur}
    \end{subfigure}
    \caption{Exploration experiments with different frontier allocation heuristic. Each colored line represents the trajectory followed by each drone, while the shaded areas symbolizes the explored regions. Colored diamonds shows drones starting points and black dots represent the obstacles in the environment.}
    \label{fig:heuristic}
    \vspace{-0.4cm}
\end{figure*}

We observed that the exploration rate, or how the area discovered evolves, increased similarly in the three scenarios as we added more drones to the exploration algorithm. Fig. \ref{fig:area_sim_results} shows the exploration results at the low obstacle density environment. Additionally, we observed that increasing the number of obstacles in the scenario drops the exploration rate (see Fig. \ref{fig:obs_dens_comp}). Fig. \ref{fig:3_final_heur} exhibits a top view of a three drones exploration on a low obstacle density universe. The top view demonstrates that the frontier allocation heuristic distributes equally the area to explore into the swarm, while other heuristics fail to.

A comprehensive summary of the results is detailed in Table \ref{tab:sim_results}. An exploration experiment is considered as completed when no frontier is available. Time starts when drones are taking off and finishes when all drones have landed. Area column represents the ratio between the known and the full map. The path length is the sum of all drone paths. Overlap stands for the intersection between all drone-explored regions. These results show (1) exploration time decreases by increasing the size of the swarm and increases by increasing the density of obstacles; (2) the area explored remains similar regardless of the number of drones; (3) path length remains similar regardless the number of drones and increases with the number of obstacles and (4) overlap increases with the number of drones and remains similar with the obstacle density.

\subsection{Real-World Experiments}
To validate our approach in real-world scenarios, we perform extensive experiments varying the swarm size, starting point of each drone, and obstacles configuration. We conduct all real-world experiments in a $8 \times 8$ $m^2$ space with an obstacle density around 0.05 $obs/m^2$. We set the navigation speed to $0.75$ $m/s$ and the spin speed to $0.1$ $rad/s$. The multi-ranger deck's maximum range was clipped up to 2m due to the available size area limitation. Complete set of parameters are shown in Table \ref{tab:real_conf}.

\begin{table}[H]
    \centering
    \begin{tabular}{c|c}
        \hline \hline
        Parameter & Value \\
        \hline
        Map resolution & 0.1 \\
        Sensor range & 2 \\
        Safety distance to obstacles & 0.25 \\
        Safety distance to other drones & 1.0 \\
        Navigation speed & 0.75 \\
        Reached distance threshold & 0.2 \\
        Spin speed & 0.1 \\
        Yaw threshold & 0.15 \\
        Minimum frontier size & 1.5 \\
        Maximum frontier size & 3.5 \\
        \hline \hline
    \end{tabular}
    \caption{Configuration for real-world experiments.}
    \label{tab:real_conf}
    \vspace{-0.3cm}
\end{table}

\begin{figure*}[!t]
    \centering
    \begin{subfigure}[b]{0.32\textwidth}
        \centering
        \includegraphics[width=\textwidth]{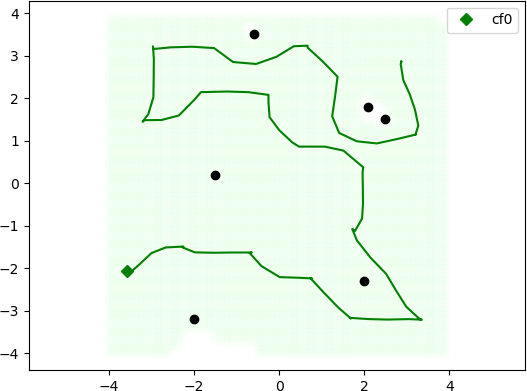}
        \caption{One drone. Time 194.0$s$ and path length 40.64$m$.}
        \label{fig:one_zv}
    \end{subfigure}
    \hfill
    \begin{subfigure}[b]{0.32\textwidth}
        \centering
        \includegraphics[width=\textwidth]{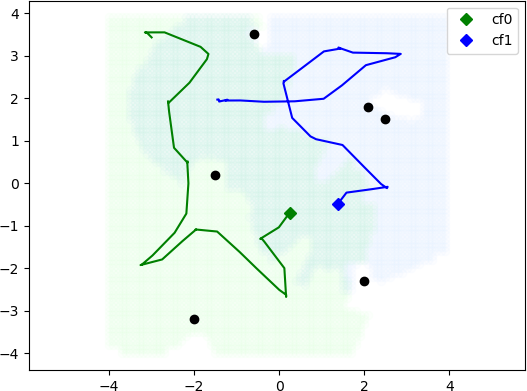}
        \caption{Two drones. Time 75.98$s$, path length 37.38$m$ and overlap 28.16$\%$.}
        \label{fig:two_zv}
    \end{subfigure}
    \hfill
    \begin{subfigure}[b]{0.32\textwidth}
        \centering
        \includegraphics[width=\textwidth]{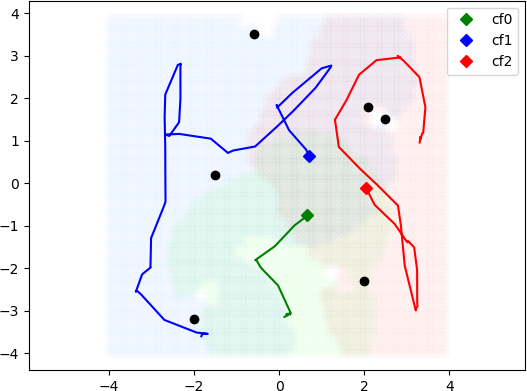}
        \caption{Three drones. Time 93.98$s$, path length 47.22$m$ and overlap 37.27$\%$.}
        \label{fig:three_zv}
    \end{subfigure}
    \caption{Real flight exploration experiments with different number of drones. Each colored line represents the trajectory followed by each drone, while the shaded areas symbolizes the explored regions. Colored diamonds shows drones starting points and black dots represent the obstacles in the environment.}
    \label{fig:real_zenithal_view}
    \vspace{-0.25cm}
\end{figure*}

Fig. \ref{fig:real_zenithal_view} shows three exploration experiments with different numbers of drones from a top view. Exploration algorithm can adapt to various swarm sizes successfully completing the mission.

\begin{figure}[H]
    \centering
    \includegraphics[width=0.43\textwidth]{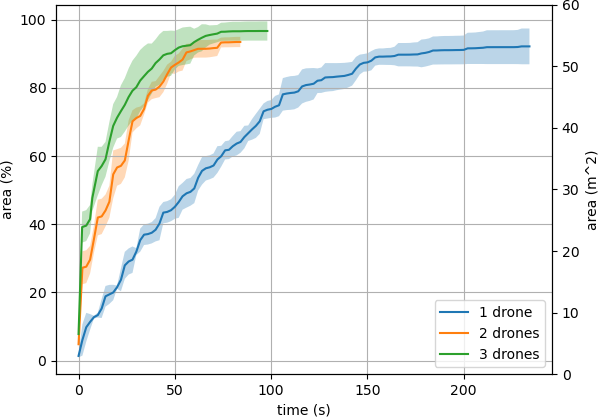}
    \caption{Average exploration rate for the real world experiments. The shaded area shows the standard deviation.}
    \label{fig:real_results}
    \vspace{-0.3cm}
\end{figure}

Fig. \ref{fig:real_results} illustrates how the area discovered evolves over time. As expected, exploration rate increases with the number of drones. Comprehensive results for real-world experiments are summarised in Table \ref{tab:real_results}. Each column description (Time, Area, Path length, Overlap) is same as that in the simulation experiments. 

\begin{table}[H]
    \centering
    \begin{tabular}{ccccc}
        \hline \hline
        \# & Time [s] & Area [\%] & Path Length [m] & Overlap [\%] \\
        \hline
        1 & 176.0 $\pm$ 31.13 & 92.16 $\pm$ 5.2 & 39.29 $\pm$ 8.48 & - \\
        2 & \textbf{78.63 $\pm$ 3.77} & 93.45 $\pm$ 1.5 & \textbf{36.9 $\pm$ 2.42} & \textbf{30.69 $\pm$ 2.38} \\
        3 & 79.25 $\pm$ 13.46 & \textbf{96.67 $\pm$ 2.8} & 47.55 $\pm$ 11.5 & 39.23 $\pm$ 11.0 \\
        \hline \hline
    \end{tabular}
    \caption{Results of the real world experiments according to the number of drones. We report the average and standard deviation for the completion time, percentage of area explored and total path traveled for the swarm. Best results are highlighted in bold.}
    \vspace{-0.2cm}
    \label{tab:real_results}
\end{table}

Results show high similarity to the simulated ones. Lowest time, path length and overlap correspond to the two drones experiment. This result is due to the size of the world, small for a three drones exploration, where the overlap increases considerably, and therefore the path length, while the time remains similar.

\begin{figure*}
    \centering
    \begin{subfigure}[b]{0.245\textwidth}
        \centering
        \includegraphics[width=\textwidth]{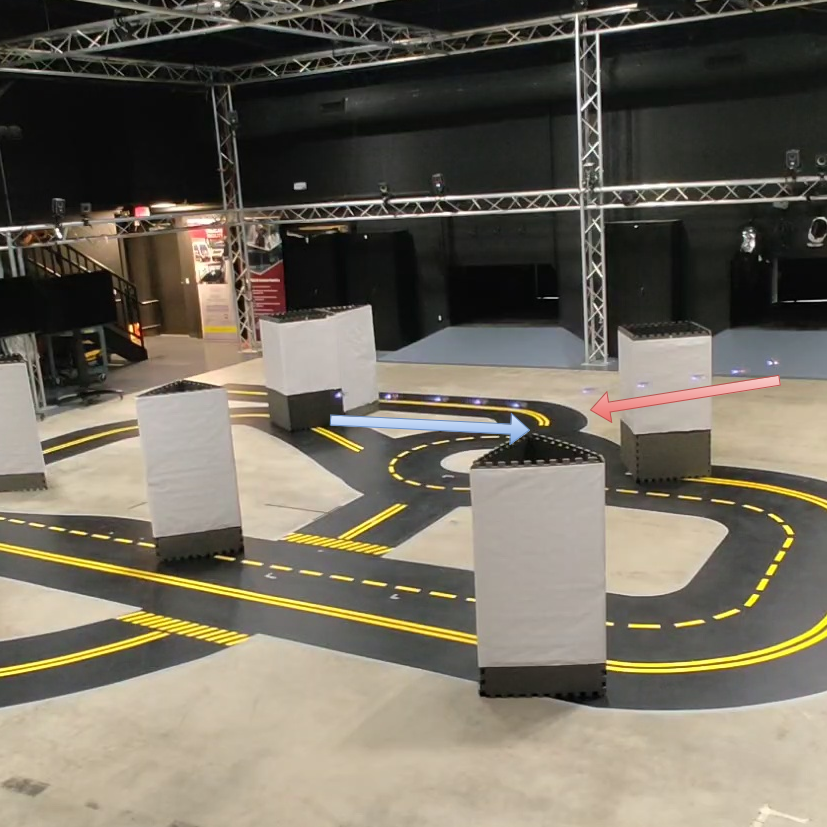}
        \caption{}
        \label{fig:coav1}
    \end{subfigure}
    \hfill
    \begin{subfigure}[b]{0.245\textwidth}
        \centering
        \includegraphics[width=\textwidth]{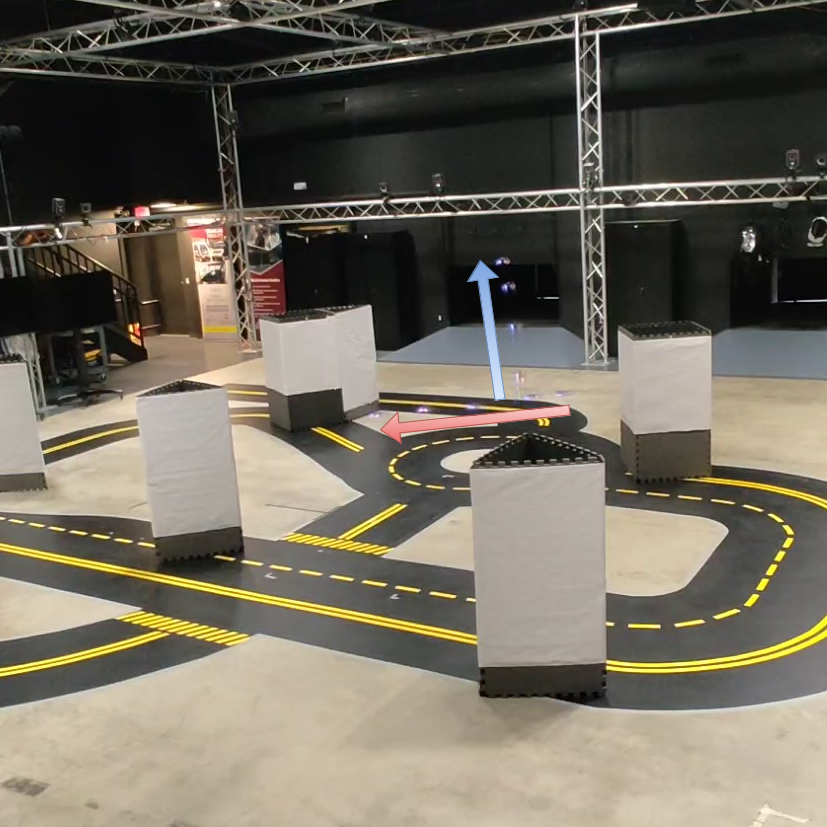}
        \caption{}
        \label{fig:coav2}
    \end{subfigure}
    \hfill
    \begin{subfigure}[b]{0.245\textwidth}
        \centering
        \includegraphics[width=\textwidth]{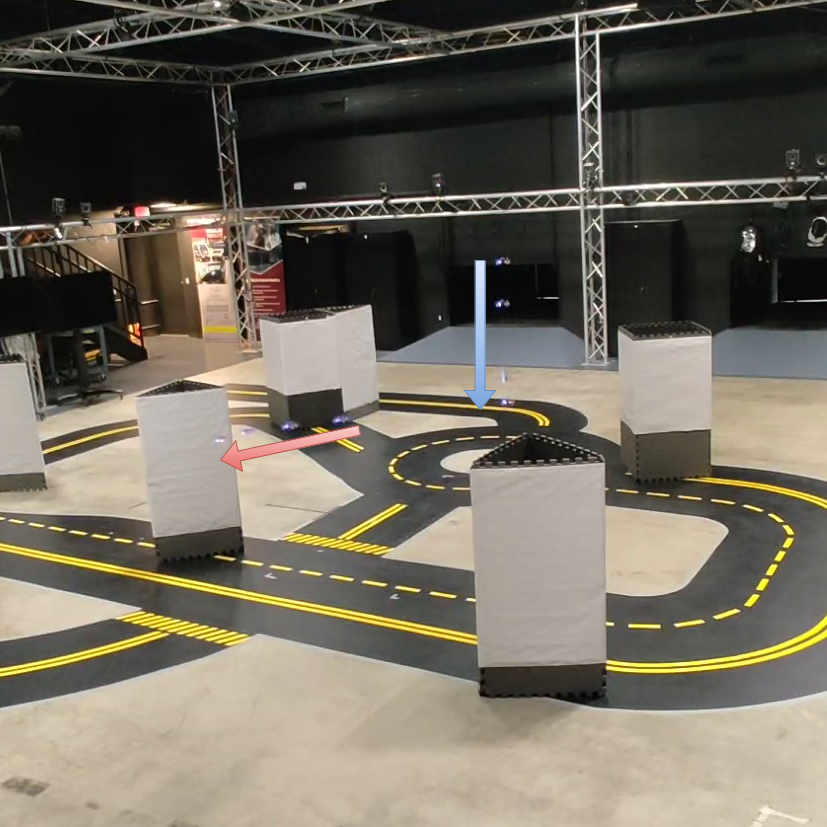}
        \caption{}
        \label{fig:coav3}
    \end{subfigure}
    \hfill
    \begin{subfigure}[b]{0.245\textwidth}
        \centering
        \includegraphics[width=\textwidth]{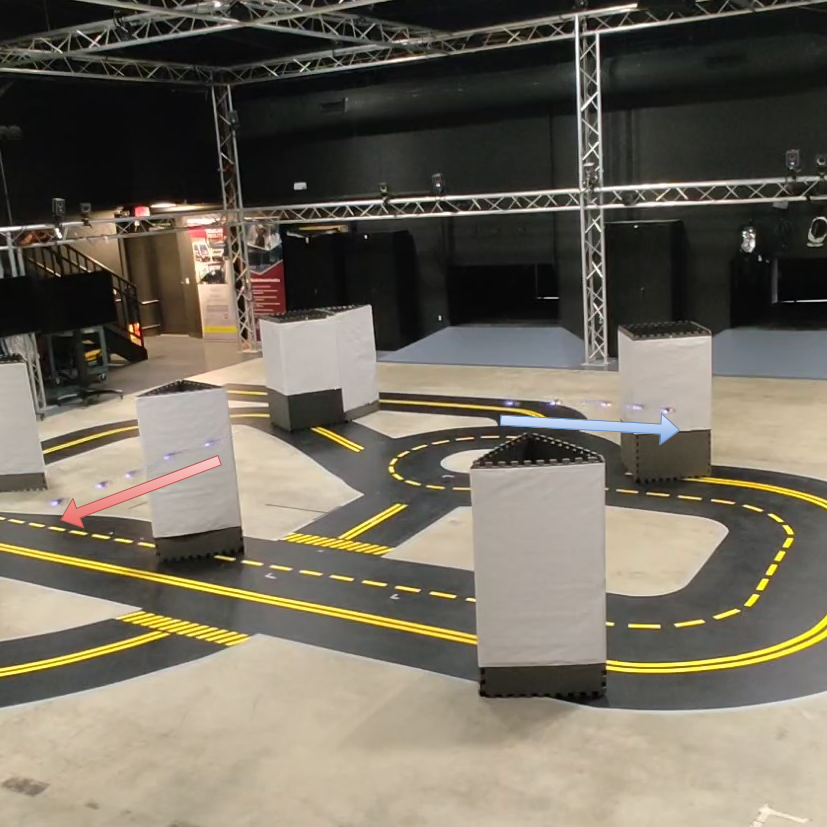}
        \caption{}
        \label{fig:coav4}
    \end{subfigure}
    \caption{Intra-swarm collision avoidance sequence. (a) A potential crash is detected; (b) drone navigator at the left is paused and the safety maneuver is started; (c) safety maneuver ends and (d) the navigator resumes its operation.}
    \label{fig:collision_avoidance}
    \vspace{-0.3cm}
\end{figure*}

Fig. \ref{fig:collision_avoidance} shows an intra-swarm collision avoidance sequence. A potential crash is detected (Fig. \ref{fig:coav1}), the drone navigator at the left is paused and a safety maneuver is started, performing a change in altitude to avoid the second drone (Fig. \ref{fig:coav2}). Once the safety distance is reestablished, the safety maneuver ends (Fig. \ref{fig:coav3}) and the navigator resumes its operation (Fig. \ref{fig:coav4}), ensuring the exploration continues safely.



\subsection{Discussion}
\subsubsection{Swarm size}
Results demonstrated that the largest swarm size achieves the fastest times thanks to our method, which evenly distributes the exploration area. However, the highest drone number are not the shortest in path due to the overlap induced. Swarm optimal size energetically speaking (roughly calculated from path length) depends on the world size and obstacle density, showing for the simulation experiments an optical number of three drones and for the real flights an optimal number of two drones. 

\subsubsection{Agent workload}
The proposed heuristic distributes equally both the area explored and path traveled by each agent of the swarm. Reducing the overlap between the areas explored increases the exploration rate, so an equal distribution of the area is crucial. Table \ref{tab:distibution_results} shows the impact of each drone in the global algorithm. Notice that the results displayed refer to each drone individually, instead to the global system ones showed in the tables before. Values obtained for the standard deviation reflect a low impact for both area and path length.

\begin{table}[H]
    \centering
    \begin{tabular}{ccc}
        \hline \hline
        \# of drones & Area / Drone [\%] & Path Length / Drone [m] \\
        \hline
        1 & 98.98 $\pm$ 0.01 & 986.66 $\pm$ 24.51 \\
        2 & 62.37 $\pm$ 4.43 & 486.20 $\pm$ 17.56 \\
        3 & 39.13 $\pm$ 2.59 & 304.37 $\pm$  9.43 \\
        5 & 28.23 $\pm$ 3.07 & 184.18 $\pm$  6.51 \\
        7 & \textbf{22.17 $\pm$ 2.64} & \textbf{148.19 $\pm$ 10.97} \\
        \hline \hline
    \end{tabular}
    \caption{Impact of each drone area explored and path travelled, mean and standard deviation, on the global algorithm for the 0.05 $obs/m^2$ density simulated scenario. Best results are highlighted in bold.}
    \label{tab:distibution_results}
    \vspace{-0.3cm}
\end{table}

\subsubsection{Frontier allocation heuristic}
Based on the proposed frontier allocation heuristic, we studied the effect of each term of the optimization problem (Eq. \ref{eq:frontiers}). On one hand, cancelling the second term (which maximizing distance between agents), each drone navigates to the nearest frontiers which behaves erratically following a pure greedy heuristic. On the other hand, cancelling the first term (which ensures going to a near frontier), each navigates to furthest frontier drawing a spiral around the map center where the exploration ends. Fig. \ref{fig:heuristic} shows a top view exploration result cancelling each term of the heuristic.




\section{Conclusions and Future Work} \label{section-conclusion}
In this work, we propose a coordination algorithm for a nano-drones swarm to address the minimal-sensing exploration problem. Our exploration pipeline leverages a novel hybrid frontier range bug algorithm that effectively handles limited sensing capabilities. This strategy allows complete exploration using just four single-beam range sensors per nano drone. Additionally, safety is ensured via a simple intra-swarm collision avoidance algorithm that prevents the drones from colliding with each other, especially during an overlap of the exploration area.


We evaluated our algorithm across various scenarios, including both simulated environments and real-world settings, and varied parameters such as world size, obstacle density, and swarm size. The results confirmed successful comprehensive explorations and demonstrated the potential advantages of using swarms of nano-drones. Our frontier allocation heuristic also showed an equal distribution of explored areas and paths traveled for each agent in the swarm, therefore reducing the exploration time as the number of drones increases.


As a part of future work, we aim to decentralize our algorithm to develop a more scalable approach. Decentralization will require better environmental representations, which are crucial for sharing information among the swarm and minimizing the data transmission to reduce the memory footprint, thus enabling the algorithm to run onboard. Furthermore, we plan to test our exploration algorithm under conditions of noisy state estimations derived from onboard sensor data.



\bibliographystyle{IEEEtran}
\bibliography{bibliography}

\end{document}